\documentclass[10pt,twocolumn,letterpaper]{article}

\usepackage[pagenumbers]{cvpr} % To force page numbers, e.g. for an arXiv version

\usepackage{multirow}
\definecolor{cvprblue}{rgb}{0.21,0.49,0.74}
\usepackage[pagebackref,breaklinks,colorlinks,allcolors=cvprblue]{hyperref}
\usepackage{algorithm} 
\usepackage{algorithmic}

\def\paperID{17067} % *** Enter the Paper ID here
\def\confName{CVPR}
\def\confYear{2026}

\title{AsyncDiff: Asynchronous Timestep Conditioning for Enhanced Text-to-Image Diffusion Inference}

\author{Longhuan Xu\\
Southeast University-Monash University\\
Joint Graduate School\\
{\tt\small l.h.xu@outlook.com}
\and
Feng Yin$^{*}$\\
Southeast University\\
{\tt\small yinfeng@seu.edu.cn}
\and
Cunjian Chen$^{*}$\\
Monash University\\
{\tt\small Cunjian.Chen@monash.edu}
}

\begin{document}
\maketitle
\begingroup
\renewcommand\thefootnote{}\footnotetext{$^{*}$Corresponding authors: Cunjian Chen, Feng Yin.}
\endgroup

\begin{abstract}
Text-to-image diffusion inference typically follows synchronized schedules, where the numerical integrator advances the latent state to the same timestep at which the denoiser is conditioned. We propose an asynchronous inference mechanism that decouples these two, allowing the denoiser to be conditioned at a different, learned timestep while keeping image update schedule unchanged. A lightweight timestep prediction module (TPM), trained with Group Relative Policy Optimization (GRPO), selects a more feasible conditioning timestep based on the current state, effectively choosing a desired noise level to control image detail and textural richness. At deployment, a scaling hyper-parameter can be used to interpolate between the original and de-synchronized timesteps, enabling conservative or aggressive adjustments. To keep the study computationally affordable, we cap the inference at 15 steps for SD3.5 and 10 steps for Flux. Evaluated on Stable Diffusion 3.5 Medium and Flux.1-dev across MS-COCO 2014 and T2I-CompBench datasets, our method optimizes a composite reward that averages Image Reward, HPSv2, CLIP Score and Pick Score, and shows consistent improvement. 
\end{abstract}    
\section{Introduction}
\label{sec:intro}

Diffusion and flow-matching models are now the dominant approach for text-to-image generation \citep{esser2024scalingrectifiedflowtransformers}. A defining aspect of diffusion models is \emph{timestep conditioning}: the denoiser is queried at a scalar time that indicates current noise level. By default, inference has two parts: the numerical integrator advances the latent along a fixed grid, and the denoiser is queried at the starting time point of each step \citep{song2021scorebasedgenerativemodelingstochastic}. We consider it \textbf{synchronous inference} because image update and velocity prediction typically follow the same timestep schedule. While this timestep coupling is consistent with the training objective of diffusion models, it seems mismatched to diffusion dynamics: image latent is updated over a \emph{time interval}, which is the difference between current timestep and next timestep; but velocity prediction is obtained at a single timestep, which assumes that the prediction from diffusion model conditioned at starting noise level is accurate enough to be used throughout the whole update interval. Such assumption can be sub-optimal at inference especially when step interval is large and noise level varies a lot.

A common approach to address this problem is using log-SNR scheduling \citep{kingma2023variationaldiffusionmodels}, which aims to make the per-step change in noise level more uniform. Then why not pick a timestep somewhere else within the update interval for velocity prediction? The principal drawback is that it breaks the consistency between training objective and inference in Eqn.~\eqref{eq:flow loss} because the model is exactly trained to predict current noise of image latent. However, this can be mitigated by keeping image latent at the interval’s start while conditioning the denoiser on a pseudo-timestep chosen to better balance the change in noise level, thus termed as \textbf{asynchronous inference}. Furthermore, we argue that the conditioning timestep fed to the denoiser can serve as a post-control signal suggesting the amount of noise retained at each step. Intuitively, conditioning as if the latent were slightly ``earlier" or ``later" in the schedule changes how the model interprets the latent, thereby modulating textural richness and detail level without altering the base generator. For example, conditioning at a later (cleaner) pseudo-timestep encourages the model to treat more of the current latent as data rather than noise, resulting in extraneous content in final image.

To sum up, asynchronous inference has three notable advantages. First, it reduces the aforementioned conflict: conditioning at a pseudo-timestep inside the update interval better reflects the effective noise seen over that step. Second, it opens a new control axis for image detail independent of base model for trading detail against cleanliness. Third, it improves evaluation scores in practice and reveals the preference of some metrics. Nevertheless, choosing that pseudo-timestep is nontrivial. The evaluation metrics we care about are non-differentiable, and the best conditioning time can be dependent on image latent, predicted velocity, as well as prompt condition. Therefore, a fixed schedule or a single global scale is inadequate. Even when a learned policy suggests nearly constant pseudo-timestep selection, the promising magnitude is unknown ahead of experiments and possibly non-linear in search space. All these challenges motivate a reinforcement learning (RL) approach that optimizes trajectory-level reward by discovering an adaptive policy for asynchronous inference while keeping diffusion model frozen.

In this paper, we introduce a reinforcement learning guided asynchronous diffusion inference method for image generation. A lightweight timestep prediction module (TPM) trained with reinforcement learning selects a pseudo-timestep, not necessarily the starting point of update interval, for velocity prediction conditioning. Velocity prediction schedule and image update schedule are thus naturally de-synchronized. Note that as explained above, we change only the timestep not image latent among the inputs to diffusion model. Our method integrates seamlessly with existing schedulers like DPM-Solver \citep{zheng2023dpmsolverv3improveddiffusionode} and is controllable at deployment by interpolating between selected pseudo-timestep and reference (original) timestep. We summarize our contributions as follows:

\begin{itemize}
\item \textbf{Asynchronous diffusion inference:} We are the first to propose the de-synchronization of velocity prediction schedule and image update schedule, which brings consistent improvement on reward metrics.
\item \textbf{RL-trained timestep predictor:} We train a lightweight timestep prediction module through reinforcement learning, re-parameterizing timestep linearly and locally with a relative, stepwise scaler.
\item \textbf{Controllable re-timing:} A single interpolation hyper-parameter that governs the aggressiveness of de-synchronization and implicitly controls image detail level, sometimes even without RL guidance.
\item \textbf{Text-to-image metric failure mode exposure:} We reveal a certain high frequency noise pattern which systematically escapes detection by some commonly used evaluation metrics. 
\end{itemize}

\begin{figure*}[t]
  \centering
  \includegraphics[width=\linewidth]{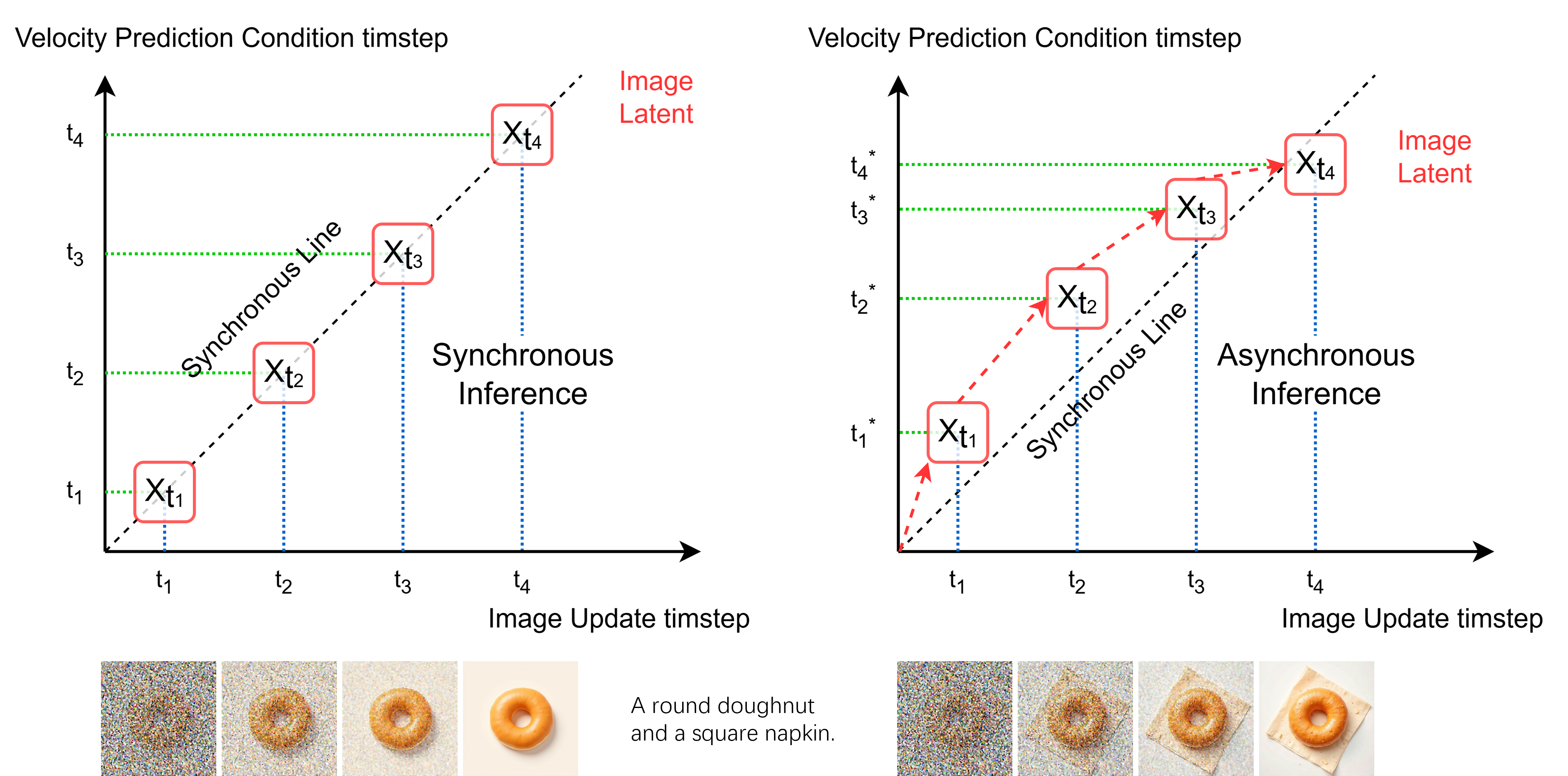}
  \caption{Asynchronous Inference.}
  \label{fig:inference}
\end{figure*}

\section{Related Works}
\label{sec:formatting}

\subsection{Flow-matching Models}
\paragraph{Rectified flows.}
Flow-matching models \citep{lipman2023flowmatchinggenerativemodeling,liu2022flowstraightfastlearning} learn the continuous normalizing flow directly, which drives the sample from Gaussian distribution into target data distribution. It achieves state of the art result with typically fewer inference steps than diffusion models \citep{ho2020denoisingdiffusionprobabilisticmodels}, and has become the dominant method of recent image generation models.

\paragraph{Inference scheduling.}
Classical diffusion samplers such as DDPM \citep{ho2020denoisingdiffusionprobabilisticmodels} and DDIM \citep{song2022denoisingdiffusionimplicitmodels} traverse a fixed grid of $T$ discrete noise levels, i.e., $t=0,1,\dots,T$, corresponding to a linear ramp of the variance parameters $\beta_t$. Flow matching models typically adopt a grid that is uniform in log signal-noise ratio (SNR) space, an empirically robust compromise that balances inference speed and image quality. Recent works \citep{ye2024trainingfreeadaptivediffusionbounded, xue2024acceleratingdiffusionsamplingoptimized, ye2025scheduleflydiffusiontime} have explored data-adaptive schedules, yet these methods keep the velocity prediction and image update schedule synchronized: they modify the grid but not the coupling. Consequently, previous works failed to explore the complicated de-synchronized schedule space where diffusion condition timestep can diverge away from image latent timestep.

\subsection{Text-to-Image Evaluation}
Evaluating text-to-image generation involves two complementary goals: (i) \emph{text to image alignment}: how faithfully an image reflects the semantics and style of the prompt, and (ii) \emph{prompt-agnostic perceptual quality}. A variety of neural network based metrics have been developed. CLIP score \citep{radford2021learningtransferablevisualmodels} was trained with large scale contrastive learning; HPS \citep{wu2023humanpreferencescorev2}, ImageReward \citep{xu2023imagerewardlearningevaluatinghuman} and PickScore \citep{kirstain2023pickapicopendatasetuser} are explicitly tuned on extensive human-preference data and feedback.

However, our experiments suggest that some existing metrics are still imperfect: they tend to reward increased local detail and structure but can be relatively insensitive to subtle high-frequency artifacts or low-amplitude Gaussian background noise.

\subsection{Reinforcement Learning}
RL has emerged as a powerful tool for steering large generative models toward non‑differentiable goals such as human preference, safety, or sampling efficiency. For text-to-image generation, the sampler can be cast as an \emph{agent} that takes action (in our case choosing the next timestep) given current state including image latent, predicted velocity, step index and prompt embedding, and receives a scalar reward once a clean image is rendered.

\paragraph{PPO.}
Proximal Policy Optimization (PPO) \citep{schulman2017proximalpolicyoptimizationalgorithms} is one of the workhorse algorithm in reinforcement learning. It stabilizes training and improves final performance by simply clipping the original objective.
\section{Proposed Method}

In this section, we first briefly go through how images are normally sampled from flow matching T2I diffusion models. Second, we explain the core idea behind our RL-trained sampler: timestep de-synchronization of image update and velocity prediction. Finally, we provide details of the implementation. 

\begin{algorithm}[t]
\caption{Default flow‑matching sampler}\label{alg:default}
\textbf{Input:} prompt $p$, steps $K$, schedule $\mathcal{T}$, guidance scale $\omega$\\
\textbf{Output:} image $\mathbf{y}$
\begin{algorithmic}[1]
\STATE $\mathbf{c}\!\leftarrow\!\text{TextEncoder}(p)$
\STATE $\mathbf{x}_{t_{0}}\!\sim\!\mathcal{N}(\mathbf{0},\mathbf{I})$
\FOR{$k=0$ \textbf{to} $K-1$}
    \STATE $\mathbf{v}_{\text{cond}}\!\leftarrow\!f_{\theta}(\mathbf{x}_{t_{k}},t_{k},\mathbf{c})$
    \STATE $\mathbf{v}_{\text{uncond}}\!\leftarrow\!f_{\theta}(\mathbf{x}_{t_{k}},t_{k},\mathbf{0})$
    \STATE $\mathbf{v_k}\!\leftarrow\!\mathbf{v}_{\text{uncond}}+\omega(\mathbf{v}_{\text{cond}}-\mathbf{v}_{\text{uncond}})$
    \STATE $\mathbf{x}_{t_{k+1}}\!\leftarrow\!\mathbf{x}_{t_{k}}+(t_{k+1} - t_k)\,\mathbf{v_k}$
\ENDFOR
\STATE $\mathbf{y}\!\leftarrow\!\text{VAEDecoder}(\mathbf{x}_{t_{K}})$
\STATE \textbf{return} $\mathbf{y}$
\end{algorithmic}
\end{algorithm}

\subsection{Preliminary}

\paragraph{Diffusion inference for image generation.}
Classic diffusion models define a forward process that gradually adds Gaussian noise to data with a predefined time-dependent variance schedule $\beta_t$ \citep{ho2020denoisingdiffusionprobabilisticmodels,song2021scorebasedgenerativemodelingstochastic}. Let $\alpha_t=1-\beta_t$ and $\bar\alpha_t=\prod_{i=1}^t \alpha_i$:
\begin{equation}
\mathbf{x}_t
= \sqrt{\bar\alpha_t}\,\mathbf{x}_0
\;+\;
\sqrt{1-\bar\alpha_t}\,\boldsymbol{\epsilon},
\qquad
\boldsymbol{\epsilon}\sim\mathcal{N}(\mathbf{0},\mathbf{I}).
\end{equation}
Flow-matching/rectified-flow models directly learn the velocity field $v_t$ defined as the time derivative of image latent\footnote{We work in the VAE latent space; the final RGB image is recovered with the pretrained decoder.} and integrate a deterministic ODE \citep{lipman2023flowmatchinggenerativemodeling} at inference based on a monotonically--decreasing time grid $\mathcal{T}=\{t_{0}=1>t_{1}>\dots>t_{K}=0\}$:  
\begin{equation}
v_t =\frac{\mathrm{d}\mathbf{x}_{t}}{\mathrm{d}t}=f_{\theta}(\mathbf{x}_{t},t,\mathbf{c}),
\quad
t = t_k,\quad k=0,1,\dots,K.
\label{eq:flow prediction}
\end{equation}
where $f_{\theta}$ predicts the \emph{velocity} (or instantaneous drift) conditioned on current image latent $\mathbf{x}_{t}$, timestep $t_{k}$ and text embedding $\mathbf{c}$. Denoting $\epsilon$ as gaussian noise, velocity training follow flow-matching loss:
\begin{equation}
\mathcal{L}
=\mathbb{E}\Big[
\big\|
\mathbf{v}_t(\mathbf{x}_t,t,\mathbf{c})
- (\boldsymbol{\epsilon}-\mathbf{x}_0)
\big\|_2^2
\Big]
\label{eq:flow loss}
\end{equation}
With an explicit Euler discretization, during inference one image update step reads:
\begin{equation}
\mathbf{x}_{t_{k+1}}
   = \mathbf{x}_{t_{k}}
   + \Delta t_{k}\,
     f_{\theta}\!\bigl(\mathbf{x}_{t_{k}},t_{k},\mathbf{c}\bigr),
\quad
\Delta t_{k}=t_{k+1}-t_{k}.
\label{eq:image update}
\end{equation}
The same timestep $t_{k}$ is used both for velocity prediction and image update which is why we call this classic method \textbf{synchronous inference}.

\paragraph{Classifier‑free guidance.}
For better text–image alignment, the sampler may employ
classifier‑free guidance (CFG) \citep{ho2022classifierfreediffusionguidance}. Two velocity predictions are performed
per step,
$f_{\theta}^{\text{cond}}$ using text condition $\mathbf{c}$ and
$f_{\theta}^{\text{uncond}}$ with the null embedding,
and are linearly combined as:
\begin{equation}
\!\!v_{\theta}^{\text{CFG}}
     = v_{\theta}^{\text{uncond}}
       + \omega\bigl(v_{\theta}^{\text{cond}}-v_{\theta}^{\text{uncond}}\bigr).
\label{eq:cfg}
\end{equation}
Here, $v_{\theta}^{\text{cond}}$ and $v_{\theta}^{\text{uncond}}$ denote velocity predictions with and without text conditioning, $v_{\theta}^{\text{CFG}}$ is the guided velocity actually used in image update, and $\omega$ is the guidance scale. 

\paragraph{Timestep selection.}
In diffusion (and flow) samplers, timestep selection means choosing a timestep grid at which the velocity field is evaluated to numerically simulate the reversed sampling process. It is a discretization schedule of a fixed number of function evaluations (NFE). In applications where fewer steps are favored, the choice of timestep can become the dominant factor for stability and fidelity at inference time. Classical inference timestep schedules include model-agnostic uniform linear schedule and log–SNR schedule, typically paired with high-order samplers such as DPM-Solver \citep{zheng2023dpmsolverv3improveddiffusionode} and UniPC \citep{zhao2023unipcunifiedpredictorcorrectorframework}. Recent works not only optimize the grid conditioned on the pretrained model \citep{tong2025learningdiscretizedenoisingdiffusion} but also make it online-adaptive to the data \citep{ye2025scheduleflydiffusiontime}.

Our asynchronous inference likewise selects timestep conditioned on both diffusion model and input data using a timestep prediction module (TPM) which iteratively outputs next timestep prediction.

\begin{algorithm}[t]
\caption{Asynchronous Sampler (TPM)}
\label{alg:async}
\textbf{Input:} prompt $p$, max steps $K_{\max}$, $\sigma_{\min}$, guidance scale $\omega$\\
\textbf{Output:} image $\mathbf{y}$
\begin{algorithmic}[1]
\STATE $\mathbf{c}\!\leftarrow\!\text{TextEncoder}(p)$
\STATE Initialize $\mathbf{x}_{t_0}\!\sim\!\mathcal{N}(\mathbf{0},\mathbf{I})$, \; $t^{\ast}_0=t_0=1$
\FOR{$k=0$ \textbf{to} $K_{\max}-1$}
    \STATE $\mathbf{v}_{\text{cond}}\!\leftarrow\!f_{\theta}(\mathbf{x}_{t_{k}},t_{k}^{\ast},\mathbf{c})$
    \STATE $\mathbf{v}_{\text{uncond}}\!\leftarrow\!f_{\theta}(\mathbf{x}_{t_{k}},t_{k}^{\ast},\mathbf{0})$
    \STATE $\mathbf{v_k}^{\ast}\!\leftarrow\!\mathbf{v}_{\text{uncond}}+\omega(\mathbf{v}_{\text{cond}}-\mathbf{v}_{\text{uncond}})$
    \STATE $\mathbf{x}^{k}_0\!\leftarrow\!\mathbf{x}_k - t_k\,\mathbf{v}_{k}^{\ast}$
    \STATE $(\alpha^{\ast}_{k+1},\beta^{\ast}_{k+1})
           \leftarrow \mathrm{TPM}_{\theta}(\mathbf{x}_{t_k},\mathbf{v^{\ast}}_k,t^{\ast}_k,\mathbf{x}^{k}_0,\mathbf{c}, k)$
    \STATE $r^{\ast}_k\!\sim\!\mathrm{Beta}(\alpha^{\ast}_k,\beta^{\ast}_k)$
    \STATE $t^{\ast}_{k+1}\!\leftarrow\!\max(\sigma_{\min},\,t_k + (0.5 +\,r^{\ast}_k)\,(t_{k+1}-t_k))$
    \STATE $\mathbf{x}_{t_{k+1}}\!\leftarrow\!\mathbf{x}_{t_{k}}+(t_{k+1} - t_k)\,\mathbf{v_k}^{\ast}$
\ENDFOR
\STATE $\mathbf{x}\!\leftarrow\!\mathbf{x}_{t_{k+1}}-t_{k+1}\mathbf{v}_{k+1}^{\ast}$
\STATE $\mathbf{y}\!\leftarrow\!\mathrm{VAEDecoder}(\mathbf{x})$
\RETURN $\mathbf{y}$
\end{algorithmic}
\end{algorithm}

\subsection{Asynchronous Inference}

\paragraph{Decoupling \emph{image update} and \emph{velocity prediction}.} 
To relax the coupling between image update schedule and velocity prediction schedule in Eqn.~(\ref{eq:flow prediction}) and Eqn.~(\ref{eq:image update}), we let the sampler query velocity field conditioned at a \emph{velocity-prediction timestep} $t^{\ast}_{k}$ that can differ from \emph{latent–update timestep} $t_{k}$ which is used to advance the sample.  Since it is infeasible to enumerate all possible timestep combinations, our goal is to find a general de-synchronized scheduling policy during sampling to yield high quality images. Concretely, learnable velocity prediction pseudo-timestep $t^{\ast}_{k}$ is parameterized as:
\begin{equation}
t_k^\ast = t_{k-1} + (0.5 + r_k^\ast)\, \Delta t_{k-1},
\quad
\Delta t_{k-1}=t_{k}-t_{k-1},
\label{eq:parameterization}
\end{equation}
which not only takes original schedule as reference baseline but also has linear and stepwise deviation scaling. The multiplicative ratio $r_k^\ast$ is drawn from beta distribution, $r^{\ast}_k\sim\mathrm{Beta}(\alpha^{\ast}_k,\beta^{\ast}_k)$, defined by the output of timestep prediction module (TPM):
\begin{equation}
%r^{\ast}_k\sim\mathrm{Beta}(\alpha^{\ast}_k,\beta^{\ast}_k),
%\quad
(\alpha^{\ast},\beta^{\ast}) = \text{TPM}_{\theta}(x_{k-1},v^{\ast}_{k-1},t^{\ast}_{k-1},x_{0}^{k-1},\mathbf{c}, k).
% \quad r^{\ast}_k\in(0,1)
% \quad \alpha_k^\ast>1,\ \beta_k^\ast>1
\label{eq:TPM}
\end{equation}
As described in Eqn.~(\ref{eq:TPM}), TPM takes several inputs: previous latent image $\mathbf{x}_{k-1}$, previous velocity $\mathbf{v^\ast}_{k-1}$, text (prompt) condition $\mathbf{c}$, calculated clean image $\mathbf{x}_{0}^{k-1}$, previous pseudo-timestep $t^\ast_{k-1}$ and step index $k$ indicating position along the whole inference process\footnote{Although clean image can be computed mathematically, passing it in directly relieves the network from having to reconstruct this exact transformation.}.

Remember the trick of asynchronous inference is to keep image latent untouched while presenting the denoiser with a dynamically chosen pseudo-timestep $t_k^\ast$, which implicitly sets the desired noise level and naturally desynchronizes the process. Now modifying only the timestep term in (\ref{eq:flow prediction}) gives the de-synchronized velocity:
\begin{equation}
v^{\ast}_{k}=f_{\theta}(\mathbf{x}_{t},t^{\ast}_{k},\mathbf{c})
\end{equation}
Accordingly, Eqn.~(\ref{eq:image update}) becomes:
\begin{equation}
\mathbf{x}_{t_{k+1}}
   = \mathbf{x}_{t_{k}}
   + \Delta t_{k}\,
     v^{\ast}_{k},
\quad
\Delta t_{k}=t_{k+1}-t_{k}
\end{equation}
Here $t_{k}$ and $t_{k+1}$ denote untouched image update schedule. In Eqn.~(\ref{eq:parameterization}), when $r_k^\ast=0.5$, asynchronous inference falls back to synchronous inference, and asynchronous pseudo-timestep value range relative to synchronous timestep is given by:
\begin{equation}
\eta = 0.5 + r_k^\ast
\end{equation}
Consequently, we define asynchronous deviation as:
\begin{equation}
D =  r_k^\ast - 0.5
\label{eq:deviation}
\end{equation} To enable controllable de-synchronization at deployment, we introduce the scaling of deviation:
\begin{equation}
\begin{aligned}
\eta_{scaled} = 1 + (r_k^\ast - 0.5)\times \gamma \\
D_{scaled} = (r_k^\ast - 0.5)\times \gamma
\end{aligned}
\label{eq:scaling}
\end{equation}
where $\gamma$ is the scaling hyper-parameter. Specifically, We conducted a comparative experiment where the manual bound on deviation is amplified into $[-1,1]$ by:
\begin{equation}
\eta^{\prime} = 2\times r_k^\ast,
\quad
D^{\prime} =  2\times (r_k^\ast - 0.5)
\label{eq:deviation prime}
\end{equation}
This comparative experiment is different from post-scaling Eqn.~(\ref{eq:scaling}) as it involves training TPMs again under new constraint.

\newcommand{\best}[1]{\textbf{#1}}
\newcommand{\second}[1]{\underline{#1}}

\begin{table*}[t]
\centering
\small
\setlength{\tabcolsep}{7pt}
\renewcommand{\arraystretch}{1.15}
\caption{SD3.5-Medium ($512{\times}512$), half precision (FP16), 15 steps.}
\label{tab:sd35}
\begin{tabular}{lc|c c c c|c}
\hline
Dataset & Method & ImageReward & HPSv2 & CLIP & PickScore & Mean Deviation \\
\hline
\multirow{3}{*}{MS-COCO 2014} & SD Base & 0.9043 & 0.2754 & \second{0.2683} & 22.27 & 0\\
                              & Our($0.5 + r^\ast$) & \second{0.9243} & \second{0.2766} & \best{0.2685} & \best{22.28} & +0.192\\
                              & Our($2 \times r^\ast$) & \best{0.9670} & \best{0.2798} & 0.2669 & \best{22.28} & +0.398\\
\hline
\multirow{3}{*}{T2I-CompBench} & SD Base & 0.8642 & 0.2698 & 0.2754 & 22.09 & 0\\
                               & Our($0.5 + r^\ast$) & \second{0.8969} & \second{0.2722} & 0.2754 & \best{22.11} & +0.231\\
                               & Our($2 \times r^\ast$) & \best{0.9377} & \best{0.2729} & \best{0.2757} & \second{22.10} & +0.106\\
\hline
\end{tabular}
\vspace{-0.75em}
\end{table*}

\begin{table*}
\centering
\small
\setlength{\tabcolsep}{7pt}
\renewcommand{\arraystretch}{1.15}
\caption{Flux.1-dev ($512{\times}512$), half precision (FP16), 10 steps.}
\label{tab:flux}
\begin{tabular}{lc|c c c c|c}
\hline
Dataset & Method & ImageReward & HPSv2 & CLIP & PickScore & Mean Deviation \\
\hline
\multirow{3}{*}{MS-COCO 2014} & Flux Base & 0.7791 & 0.2769 & 0.2592 & 22.60 & 0\\
                              & Ours($0.5 + r^\ast$) & \second{0.9267} & \second{0.2871} & \second{0.2600} & \best{22.74} & +0.499\\
                              & Ours($2 \times r^\ast$) & \best{0.9452} & \best{0.2920} & \best{0.2601} & \second{22.70} & +0.959\\
\hline
\multirow{3}{*}{T2I-CompBench} & Flux Base & 0.7197 & 0.2747 & 0.2655 & 22.46 & 0\\
                               & Ours($0.5 + r^\ast$) & \best{0.8725} & \second{0.2854} & \best{0.2673} & \best{22.57} & +0.499\\
                               & Ours($2 \times r^\ast$) & \second{0.8584} & \best{0.2866} & \second{0.2667} & \second{22.52} & +0.640\\
\hline
\end{tabular}
\vspace{-0.75em}
\end{table*}

\paragraph{Trajectory optimization.}
We train TPM with \emph{Group‑Relative PPO} (GRPO) \citep{shao2024deepseekmathpushinglimitsmathematical} which, for each prompt, generates multiple samples and computes normalized advantages using the group mean and variance. Compared with PPO:
\begin{equation}
\mathcal{L}_{\text{PPO}}
=\mathbb{E}\!\Bigl[
\min\!\bigl(r\,\hat A,\;
\text{clip}(r,1-\varepsilon,1+\varepsilon)\,\hat A\bigr)
\Bigr],
\quad
r=\frac{\pi_{\theta}(\tau)}{\pi_{\text{old}}(\tau)}.
\end{equation}
GRPO removes the need for an explicit value network while preserving the variance-reduction benefits of a baseline. During training, we roll out a \emph{full} sampling loop until termination that alternates between frozen diffusion model and  trainable TPM. The resulting schedule $\hat{\mathcal{T}}=(t_1,\dots,t_N)$ is a complete trajectory whose log‑probability factorizes as:
\begin{equation}
\pi_\theta(\hat{\mathcal{T}})=
\prod_{i=1}^{N}
\pi_\theta^{(i)}\!\bigl(t_i \bigr)
\label{eq:traj_policy}
\end{equation}
The whole trajectory is treated as a single action, on which we do  PPO‑clip with a trajectory‑level surrogate. It  terminates when (i) a global step cap $K_{\max}=15$ (SD3.5‑medium) or $10$ (Flux.1‑dev) is reached, or
(ii) $t_{k+1}<\sigma_{\min}=10^{-3}$ for all samples in the batch. A final Euler step brings the latent to $t=0$.

\paragraph{Composite RL objective.}
Each trajectory is assigned a scalar reward that averages over four diverse metrics: Image Reward, HPSv2, CLIP Score and Pick Score. Each metric is z-score normalized within the batch to ensure numerical balance.
\begin{equation}
R = \tfrac14 \sum_{i\in\{\text{IR},\,\text{HPS},\,\text{CLIP},\,\text{Pick}\}} \hat{S}_{i},
\qquad
\hat{S}_{i}=\frac{\mathrm{Score}_{i}-\mu_i}{\sigma_i+\varepsilon}
\end{equation}

\begin{figure*}[t]
  \centering
  % tighten vertical gaps for this figure only (optional)
  % \captionsetup[sub]{aboveskip=2pt, belowskip=2pt}

  % \begin{subfigure}[t]{0.9\linewidth}
  %   \centering
  %   \includegraphics[width=\linewidth]{sec/Figures/Scaling1.png}
  % \end{subfigure}\vspace{2pt}

  % \begin{subfigure}[t]{0.9\linewidth}
  %   \centering
  %   \includegraphics[width=\linewidth]{sec/Figures/Scaling2.png}
  % \end{subfigure}\vspace{2pt}
  \includegraphics[width=0.9\linewidth]{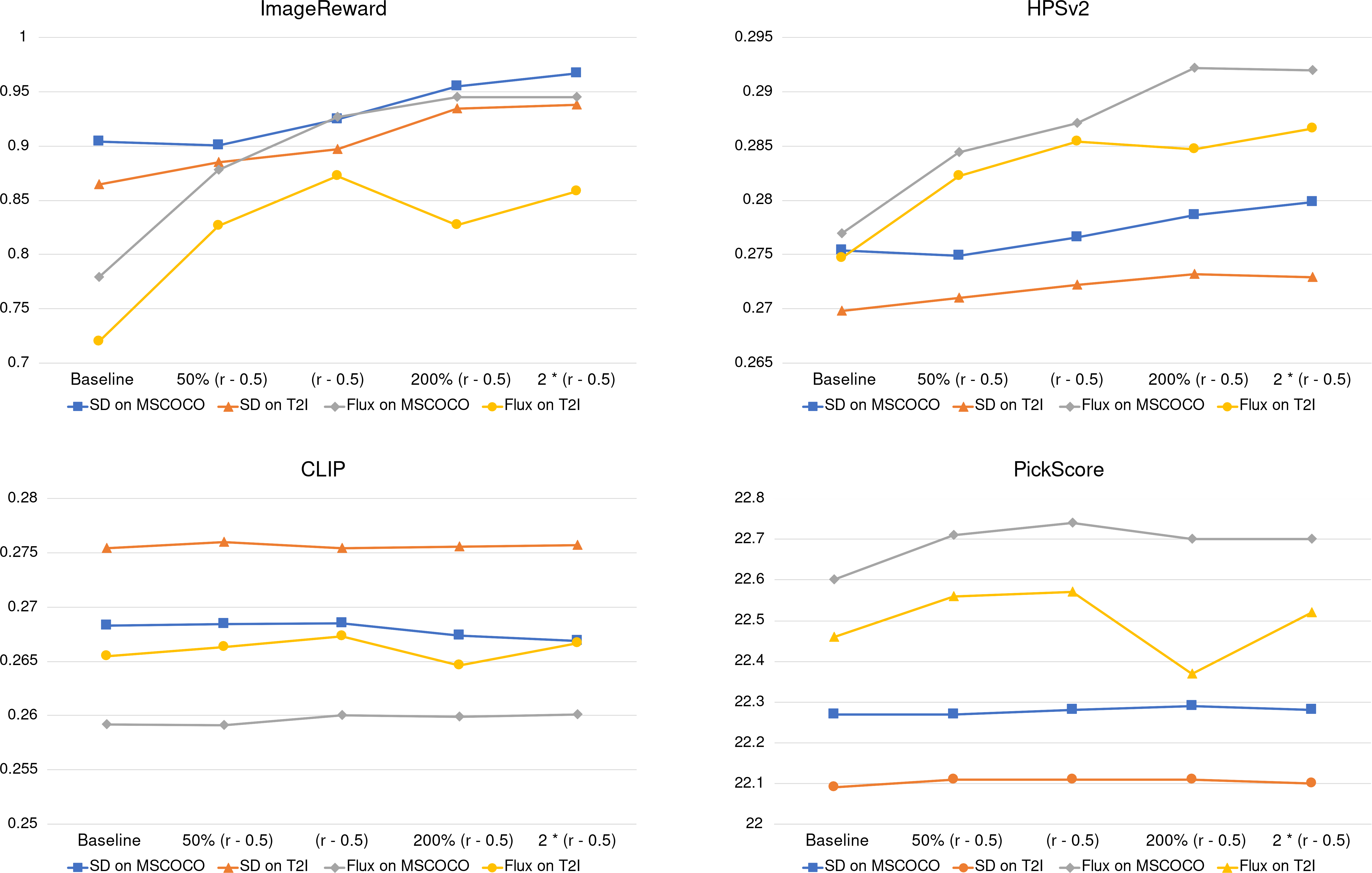}
  \caption{Deviation scaling result. Each chart plots evaluation metrics as we scale the per-step deviation $(r_k^\ast-0.5)$ from (\ref{eq:deviation}) using the factor $\gamma$ in (\ref{eq:scaling}) by 50\%, 100\% and 200\%. Synchronous inference baseline and comparative experiments with lifted deviation bound (\ref{eq:deviation prime}) denoted as $ 2 * (r_k^\ast-0.5)$ are also covered. Curves include ImageReward, HPSv2, CLIP, and PickScore (all higher is better) on MS-COCO 2014 and T2I-CompBench from SD and Flux models.}
  \label{fig:scaling}
\end{figure*}

\paragraph{Timestep Prediction Module architecture.}
TPM adopts a \emph{token‑centric} design that casts model input into a short transformer sequence. Drop-out and layer-normalization are disabled to avoid unnecessary randomness or complexity and smooth training.

\begin{enumerate}
\item \textbf{Latent tokens.}
      The three spatial tensors—noisy latent, predicted
      flow and clean image—are first cut into patches and then projected into a joint embedding space.

\item \textbf{Condition tokens.}
      A handful of condition tokens are appended to the sequence, including \emph{Text‑condition tokens} obtained by pre-processing the prompt and \emph{temporal tokens} that encode image update timestep.

\item \textbf{Backbone.}
      All tokens are fed to a shallow 4-layer
      transformer encoder.

\item \textbf{Read‑out head.}
      The final set of global tokens is flattened and passed through a
      small MLP that emits two scalars
      \((a^{\ast},b^{\ast})\).
      Each scalar is mapped to a strictly positive value via
      \begin{equation}
        \phi(x)=
        \begin{cases}
          2 + x + \tfrac{1}{2}x^{2}, & x>0,\\[4pt]
          1 + e^{x}, & x\le 0,
        \end{cases}
        \label{eq:unimodal}
      \end{equation}
      producing parameters
      \((\alpha^{\ast},\beta^{\ast})\) required in Eqn.~(\ref{eq:TPM}) and ensuring a uni-modal Beta distribution.

\end{enumerate}
\section{Experimental Results}

\subsection{Experimental Setup}
\paragraph{Implementation details.}
We use publicly available Flux.1‑dev and SD 3.5‑medium text-to-image flow matching models in \textbf{half‑precision} (FP16),  generating $512\times512$ images. Gradient is normalized at 1.0 and classifier‑free guidance scale is set to 5. Inference is capped at $K_{\max}=10$ steps (Flux) or $15$ steps (SD\,3.5). We adopt PPO clip ratio $\varepsilon=0.2$ and constant learning rates: 2e-5 for Flux and a smaller 1e-5 for SD since SD undergoes more inference steps and we do trajectory-level optimization.

\paragraph{Datasets and training.}
Prompts are drawn from the MS-COCO 2014 captions \citep{lin2015microsoftcococommonobjects}
and T2I-CompBench dataset \citep{huang2025t2icompbenchenhancedcomprehensivebenchmark}, following official training-validation split. Specifically, MS-COCO 2014 provides over 400k training captions and over 200k validation captions, while T2I-CompBench offers $\sim$5.5k training prompts and $\sim$2.4k validation prompts. We fix the random seed to 42, randomly sample 2,048 MS-COCO validation prompts, and use the entire T2I-CompBench validation set. Training uses the full training splits of both datasets. In each iteration, we draw a single prompt and pair it with 16 Gaussian noise latents, then apply GRPO with a mini-batch size of 4, using the mean score over the 16 rendered images as the baseline to reduce inter-prompt variance and stabilize the advantage estimates. For both datasets and for both SD and Flux backbones, the TPM is trained for 4k iterations (i.e., 4k prompts $\times$ 16 image samples).

\paragraph{Evaluation metrics.}
We report four automatic metrics used both for training and evaluation: (i) ImageReward \citep{xu2023imagerewardlearningevaluatinghuman}, a reward model distilled from human feedback that correlates with human judgments of aesthetic quality and alignment;  (ii) HPSv2 \citep{wu2023humanpreferencescorev2}, a learned preference model trained on large-scale human comparisons that reflects perceived image quality and prompt faithfulness; (iii) CLIP Score \citep{radford2021learningtransferablevisualmodels}, the cosine similarity between CLIP image and text embeddings, which primarily measures semantic alignment between the prompt and the generated image; and (iv) PickScore \citep{kirstain2023pickapicopendatasetuser}, a preference predictor trained on the Pick-a-Pic dataset capturing overall human-perceived appeal.

\subsection{Main Results}
During testing or validation, $r_k^\ast$ is no longer a random variable, but instead a deterministic value given by:
\begin{equation}
r_k^\ast
=\underset{r\in(0,1)}{\arg\max}\; r^{\alpha_k^\ast-1}(1-r)^{\beta_k^\ast-1}
=\frac{\alpha_k^\ast-1}{\alpha_k^\ast+\beta_k^\ast-2},
\end{equation}
which is well-defined here since (\ref{eq:unimodal}) ensures $\alpha_k^\ast>1$ and $\beta_k^\ast>1$. The value of $r_k^\ast$ ranges over $[0,1]$, thus deviation is constrained within $[-0.5,0.5]$. Empirically, we notice that on Flux models the TPM frequently indeed hits the +0.5 deviation ceiling when deviation bound is not lifted (\ref{eq:deviation}), indicating that Flux is more adaptable to asynchronous timestep drift, and that our comparative experiments (\ref{eq:deviation prime}) are necessary.

Main results for synchronous inference baseline and asynchronous inference with or without lifted bound are reported in Table~\ref{tab:sd35} and Table~\ref{tab:flux}. Quantitatively, compared against default synchronized sampling, our asynchronous inference yields consistent metric gains across datasets and models. All TPMs tend to push denoise timestep forward (closer to clean image time point) as implied by positive deviation value, effectively presenting a lower noise level to the denoiser for extra detail/content. This agrees with our initial guess that choosing a denoising timestep somewhere in the middle of image update interval may be beneficial.

Nevertheless, Flux benefits much more from asynchronous inference, likely for two reasons: (i) the Flux model is more robust and better at handling excessive noise; and (ii) it uses fewer inference steps, yielding larger timestep intervals and therefore a more serious misalignment between velocity-prediction time point and image-update interval. In other words, \textbf{asynchronous inference may yield great metric gains when the total number of inference steps is small with powerful diffusion backbone models}.

We also observe clear trade-offs among the four metrics: improving one can depress another. \citet{xue2025dancegrpounleashinggrpovisual} report a similar tension—training Flux solely on HPSv2 reduces CLIP and GenEval~\citep{ghosh2023genevalobjectfocusedframeworkevaluating}. In our case, asynchronous inference with position deviation tends to increase ImageReward and HPSv2 while slightly reducing CLIP and PickScore. In comparative experiments, lifting the bound of deviation indeed achieves better ImageReward and HPSv2, but is at the risk of suffering a severe drop in CLIP and PickScore.

% \begin{figure}[t]
%   \centering
%   \includegraphics[width=\linewidth]{scaling.png}
%   \caption{Deviation scaling.}
%   \label{fig:scaling}
% \end{figure}

\begin{figure}
  \centering
  \includegraphics[width=0.9\linewidth,clip,trim=16 24 16 10]{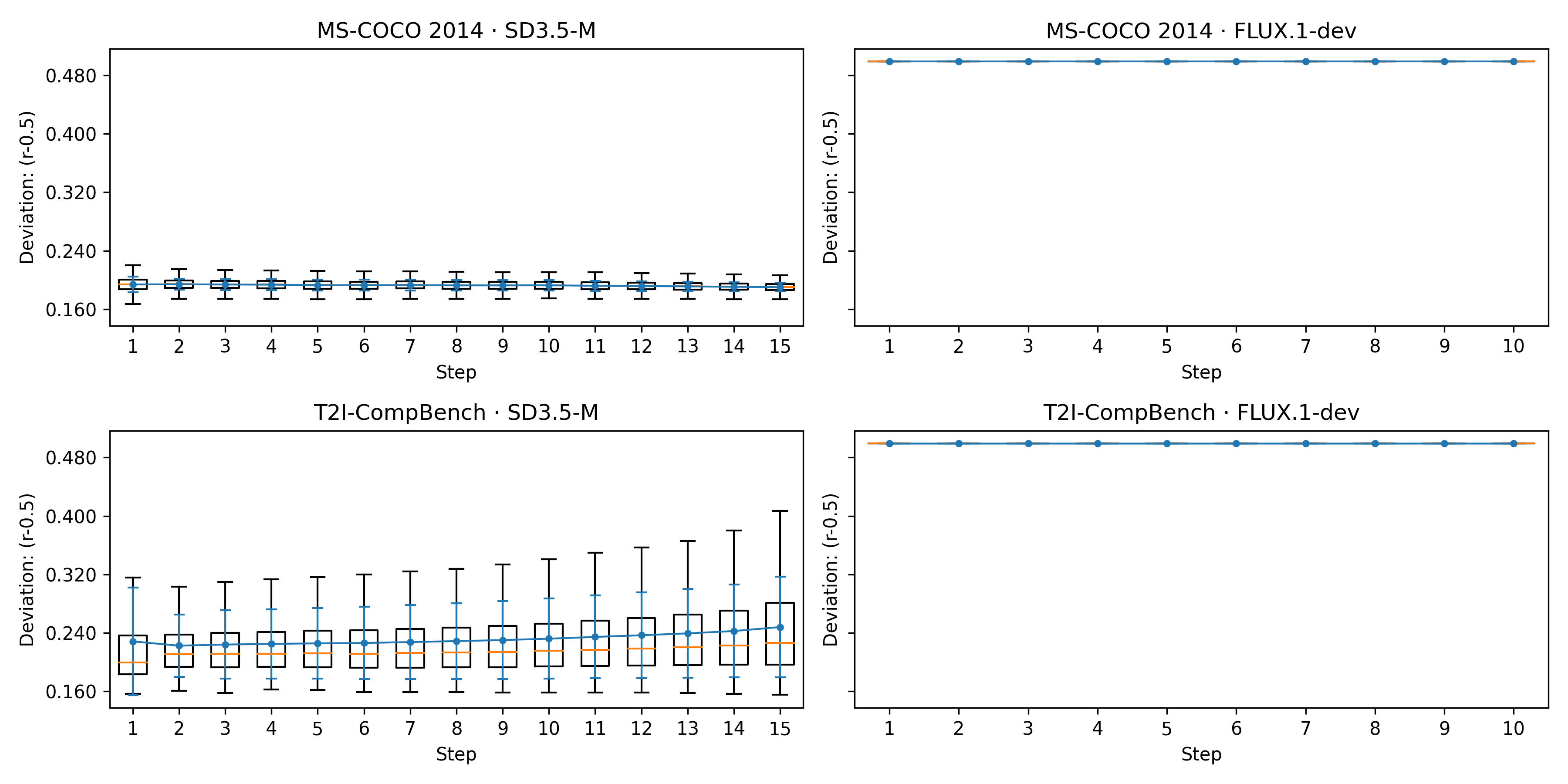}
  \caption{Per-step deviation: $(r-0.5)$ distributions across datasets/models.}
  \label{fig:trajectory1}
\end{figure}

\begin{figure}
  \centering
  \includegraphics[width=0.9\linewidth,clip,trim=16 24 16 10]{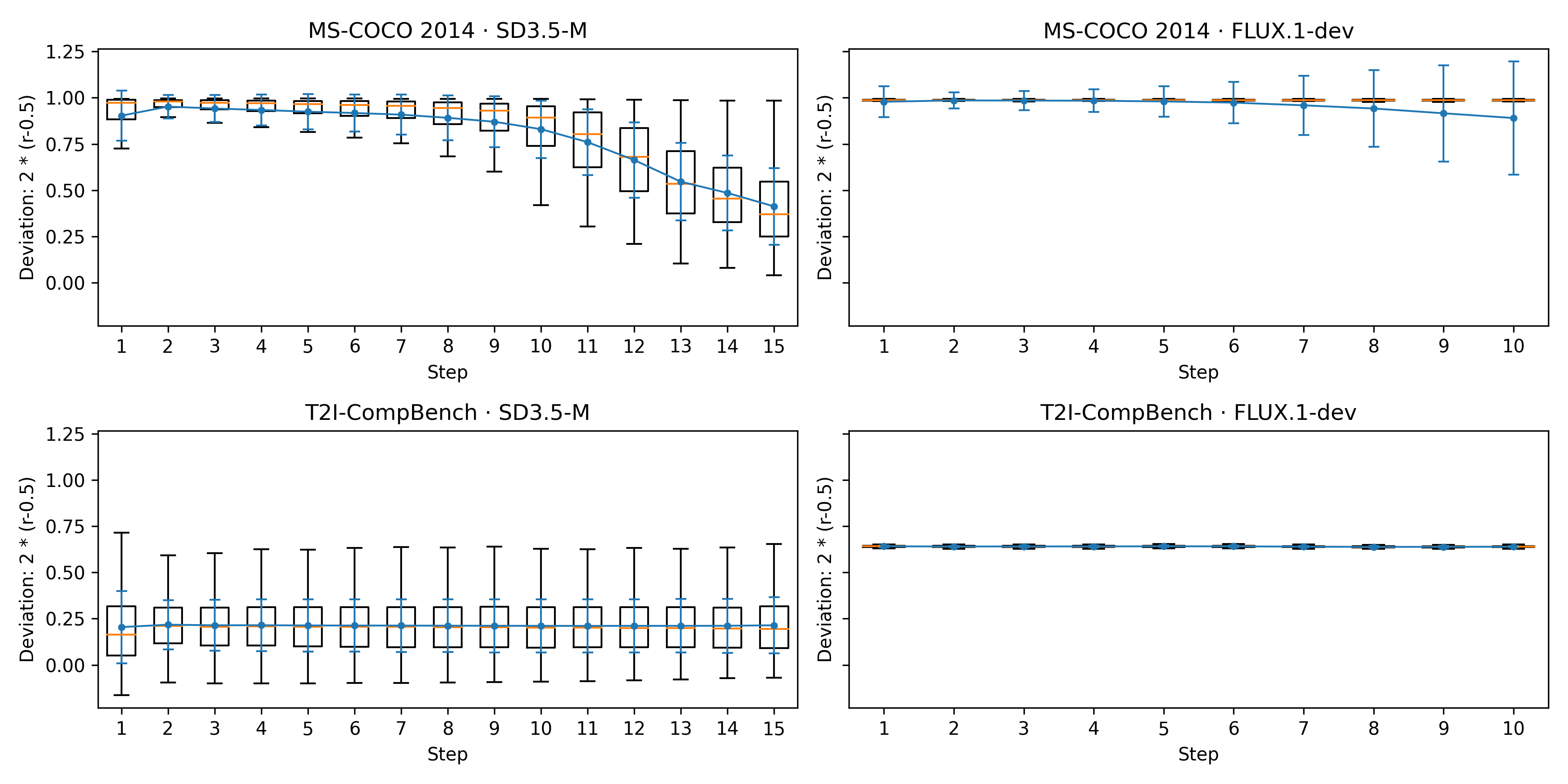}
  \caption{Per-step deviation of comparative experiments: $2\times (r-0.5)$.}
  \label{fig:trajectory2}
\end{figure}

\begin{figure}
  \centering
  \includegraphics[width=\linewidth]{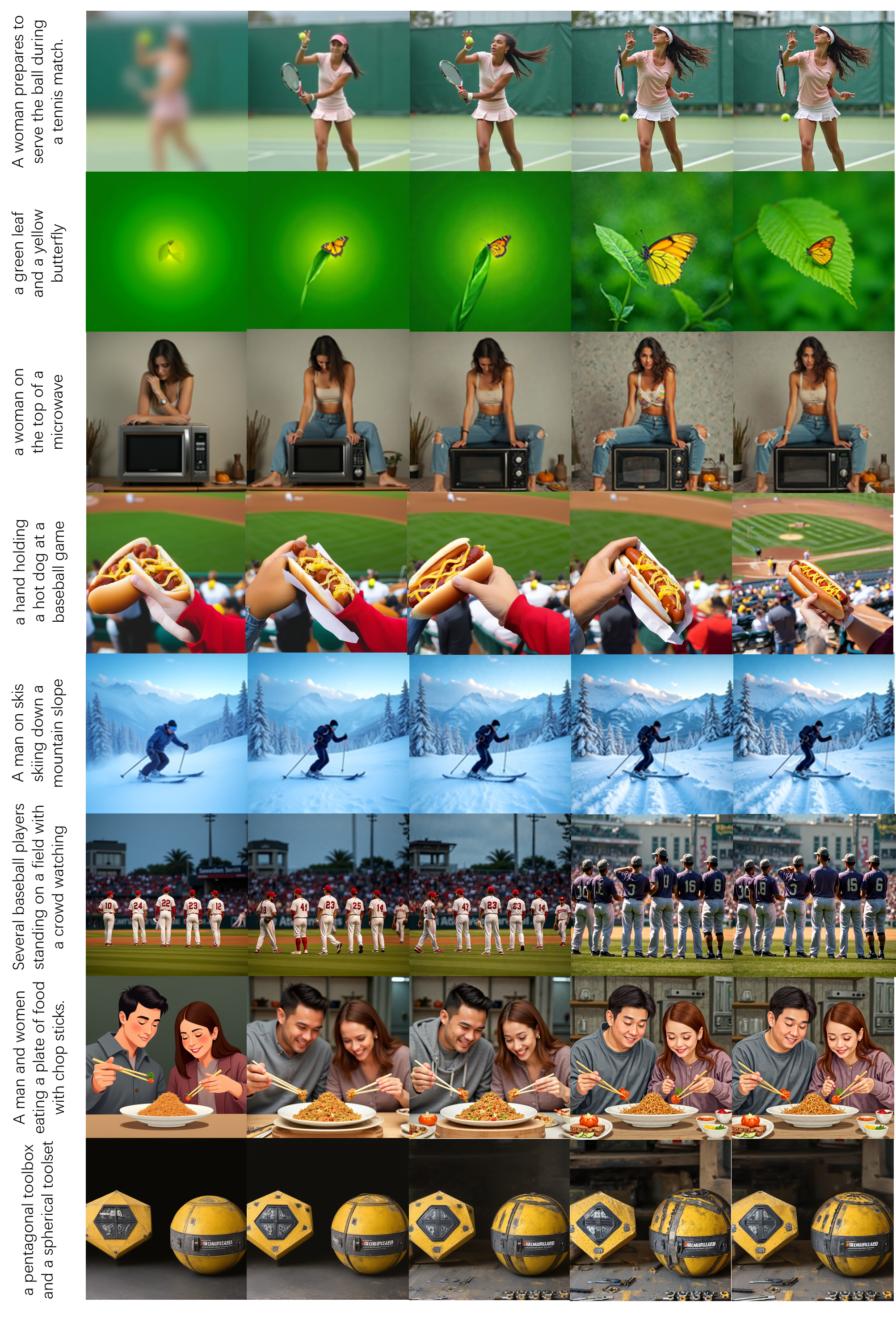}
  \caption{Qualitative results. Each image in each panel from left to right correspond to: synchronous inference (baseline); 50\%, 100\% and 200\% scaled asynchronous inference with deviation $(r_k^\ast-0.5)$; comparative asynchronous inference with deviation $2 * (r_k^\ast-0.5)$.}
  \label{fig:panel}
\end{figure}

Figs~(\ref{fig:trajectory1}) (\ref{fig:trajectory2}) show the metric results of scaling the original deviation capped within $[-0.5,0.5]$. Together with qualitative visualization in Figs~(\ref{fig:panel}), \textbf{we argue that larger (positive) deviations yield richer, more detailed—and sometimes busier—images favored by ImageReward and HPSv2, whereas CLIP and PickScore do not always benefit, because some unsolved Gaussian background noise would also be left in the image}. This underscores that different metrics emphasize different aspects of quality, and that \textbf{many metrics failed to take high frequency noisy patterns in final image into consideration} but are instead satisfied with low level details, while other metrics focus on semantic prompt alignment.

From left to right within each panel in Figs~(\ref{fig:panel}), the images correspond to: baseline (synchronized), 50\% scaled, 100\% scaled, 200\% scaled version of $(r-0.5)$ deviation (\ref{eq:scaling}), and comparative $2\times (r-0.5)$ deviation (\ref{eq:deviation prime}). In some cases like the first, second and third rows in Figs~(\ref{fig:panel}), the denoiser is allowed the chance to denoise missing objects and fix errors; in general image detail level and textural richness are increased, with the annoying drawback of leaving some unsolved Gaussian noise in background, as shown in the fourth, fifth and sixth rows in Figs~(\ref{fig:panel}). Although the metric gains can be occasionally modest, the qualitative improvement of crisper textures and richer structure is consistent and clearly visible.


Besides, the deviation of each step varies along the trajectory, and TPMs trained on different datasets or backbones adopt distinct policies Figs~(\ref{fig:trajectory1}, \ref{fig:trajectory2}): Flux shows near-constant deviation across steps with low variance, whereas SD exhibits higher variance with sometimes a characteristic pattern—large early deviations that gradually taper toward later steps. \textbf{When the per-step deviation exhibits low variance, an RL-free policy with a fixed deviation level—agnostic to sample context—may suffice and simplify deployment}.

\section{Alternative Approach and Future Work}

\begin{figure}
  \centering
  \includegraphics[width=0.9\linewidth]{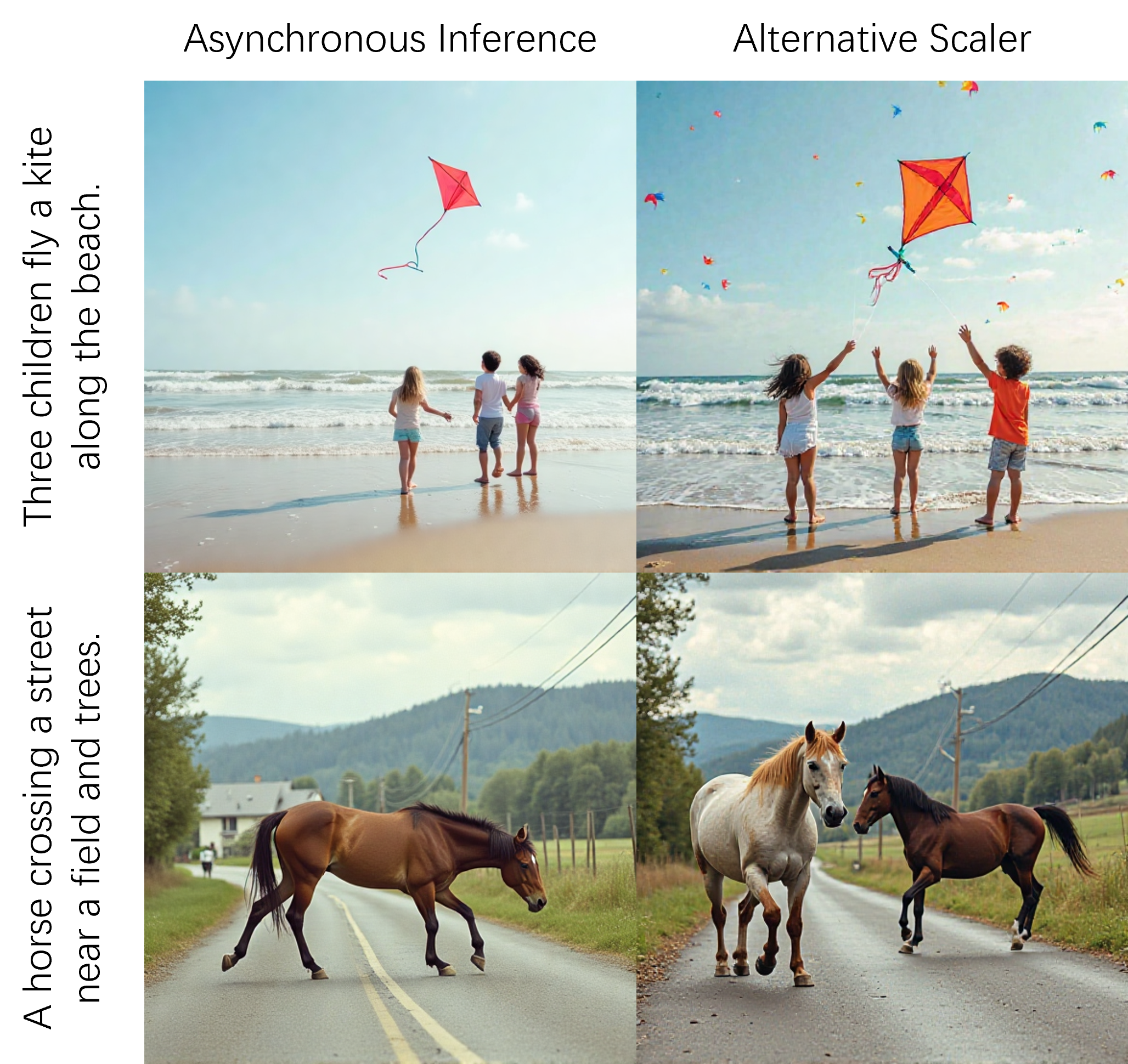}
  \caption{Qualitative comparison of proposed asynchronous inference and alternative approach.}
  \label{fig:Alternative}
\end{figure}

The idea of deliberately forcing the diffusion model to treat part of the noise as image data (and vice versa) can also be leveraged in another way. Instead of modifying the inference timestep schedule, an independent scalar $\omega$ can be multiplied into Eqn.~\eqref{eq:image update}, yielding:
\begin{equation}
\mathbf{x}_{t_{k+1}}
   = \mathbf{x}_{t_{k}}
   + \Delta t_{k}\,
     f_{\theta}\!\bigl(\mathbf{x}_{t_{k}},t_{k},\mathbf{c}\bigr)\,
     \omega
\label{eq:alternative approach}
\end{equation}
Similarly,  $\omega$ is set within $[0.5,1.5]$ by:
\begin{equation}
\omega = 0.5 + r_k^\ast,
\quad
r_k^\ast \in \left[0, 1\right] 
\end{equation}
Likewise, when $\omega$ is smaller than 1, some noise will be left to be interpreted as image data while keeping timstep untouched. Although this alternative approach appears promising, early experiments indicate that it is less stable and more prone to meaningless high-frequency patterns than asynchronous inference, as is qualitatively illustrated in Fig~(\ref{fig:Alternative}). 

\section{Conclusion}

In this paper, we introduced RL-Guided Asynchronous Diffusion Inference, a sampling method that de-synchronizes the denoiser’s conditioning timestep from the latent-update schedule. The approach is lightweight and plug-and-play, offers a single scaling hyper-parameter for controllable de-synchronization, remains compatible with standard schedulers, and delivers consistent gains across backbones and datasets—particularly on ImageReward and HPSv2—by letting the TPM present a cleaner effective noise level to the denoiser.

There are, however, limitations: RL training is not fully stable—especially for the SD model on MS-COCO—where the learned policy can diverge between lifted and unlifted deviation-bound settings and drift over training. In addition, current reward signals insufficiently penalize residual high-frequency artifacts; a more robust metric that explicitly distinguishes structured detail from spurious noise is needed as an RL objective.

In the future, we will integrate de-synchronization with a wider range of schedulers and high-order solvers, scale training with diverse inference steps and broader datasets to better map quality–efficiency trade-offs, refine the alternative independent scalar by narrowing its value range to suppress unwanted noisy patterns, and combine this multiplicative control with asynchronous timestep conditioning to exploit their complementary strengths.

{
    \small
    \bibliographystyle{ieeenat_fullname}
    \bibliography{main}
}

% WARNING: do not forget to delete the supplementary pages from your submission 
% \input{sec/X_suppl}

\end{document}